\documentclass{article}

\usepackage{PRIMEarxiv}

\usepackage[utf8]{inputenc} 
\usepackage[T1]{fontenc}    
\usepackage{hyperref}       
\usepackage{url}            
\usepackage{booktabs}       
\usepackage{amsfonts}       
\usepackage{nicefrac}       
\usepackage{microtype}      
\usepackage{lipsum}
\usepackage{fancyhdr}       
\usepackage{graphicx}       
\graphicspath{{media/}}     

\title{A Sentinel-3 foundation model for ocean colour}

\author{
 Geoffrey Dawson \\
  IBM Research Europe\\
   \And
 Remy Vandaele \\
  University of Exeter \\
  \And
 Andrew Taylor\\
  STFC Hartree Centre \\
  \And
  David Moffat \\
  Plymouth Marine Laboratory \\
  National Center for Earth Observation \\
  \And
  Helen Tamura-Wicks \\
  IBM Research Europe\\
  \And
  Sarah Jackson \\
  STFC Hartree Centre \\
  \And
  Rosie Lickorish \\
  IBM Research Europe\\
  \And
  Paolo Fraccaro \\
  IBM Research Europe\\
  \And
  Hywel Williams \\
  University of Exeter \\
  \And
  Chunbo Luo \\
  University of Exeter \\
  \And
  Anne Jones \\
  IBM Research Europe\\
}

\begin{document}
\maketitle

\begin{abstract}
Artificial Intelligence (AI) Foundation models (FMs), pre-trained on massive unlabelled datasets, have the potential to drastically change AI applications in ocean science, where labelled data are often sparse and expensive to collect. In this work, we describe a new foundation model using the Prithvi-EO Vision Transformer architecture which has been pre-trained to reconstruct data from the Sentinel-3 Ocean and Land Colour Instrument (OLCI). We evaluate the model by fine-tuning on two downstream marine earth observation tasks. We first assess model performance compared to current baseline models used to quantify chlorophyll concentration. We then evaluate the FM’s ability to refine remote sensing-based estimates of ocean primary production. Our results demonstrate the utility of self-trained FMs for marine monitoring, in particular for making use of small amounts of high quality labelled data and in capturing detailed spatial patterns of ocean colour whilst matching point observations. We conclude that this new generation of geospatial AI models has the potential to provide more robust, data-driven insights into ocean ecosystems and their role in global climate processes.
\end{abstract}

\section{Introduction}
\label{introduction}

Artificial Intelligence (AI) foundation models are being increasingly used for deep learning based Earth observation tasks \cite{zhang2024}. These models are trained on a large amount of unlabelled data using self-supervised learning, then the corresponding weights are fine-tuned using supervised learning methods for a variety of downstream tasks. One of the main advantages of a foundation model is that it can achieve equal or better performance in downstream tasks compared to other deep learning methods, with better computational efficiency and less labelled data \cite{jakubik2023}. 

Satellite remote sensing is currently the best tool for viewing the ocean surface globally and systematically at high temporal and spatial resolutions. To date, remote sensing foundation models have primarily been developed for land applications \cite{lu2024}. However, this approach could be particularly valuable for the ocean, where many applications rely on small amounts of high-quality data taken from in-situ measurements, a limitation that has been recognised for ocean remote sensing applications of deep learning \cite{Wang2024}. Whilst general remote sensing foundation models have been fine-tuned and evaluated for ocean applications such as marine object detection \cite{jakubik2025} and classification \cite{Glaser2025}, and foundation model approaches have been explored for object detection in ground and ocean-based photography \cite{Orenstein2025, Zheng2025}, the only ocean colour remote sensing foundation model published to date is (to our knowledge) is the OCFM presented by \cite{Yang2025}. OCFM is a pixel based model, pre-trained on 4~km MODIS-Aqua satellite data, using a three phase training approach: pre-training with MODIS-Aqua daily values across multiple bands, and a target of MODIS-Aqua level 3 products (derived from ocean colour algorithms); general fine-tuning for ocean color; and task-specific application fine-tuning. The model outperformed baselines and algorithmic approaches to estimation primary production and water clarity applications, but is limited by the low resolution and lack of near-infrared bands in MODIS data.   

Here we present the first vision transformer based remote sensing foundation model trained on 300~m resolution Sentinel-3 Ocean and Land Colour Instrument (OLCI) and Sea and Land Surface Temperature Radiometer (SLSTR) data over the ocean. The resulting foundation model, with subsequent fine tuning, has a range of potential use cases across Sentinel-3 OCLI applications: from tracking harmful algal blooms to monitoring estuarine processes. Here, we evaluate the model by fine-tuning for two downstream applications: quantification of chlorophyll-a concentration, and quantification of ocean primary production.

Chlorophyll-a concentration is a key variable measured by Sentinel-3: it is fundamental in our ability to characterise marine biomass and is clasified by the European Space Agency as an ECV (essential climate variable). It is well studied by ocean remote sensing (e.g. \cite{TILSTONE2021, Tran2023, VANHELLEMONT2021}), and therefore provides the opportunity to measure the performance of the foundation model on a well known application. There have been several approaches to derive chlorophyll-a using deep learning as outlined in the review by \cite{yuan2020}. Additionally, there are two chlorophyll-a concentrations supplied in the Sentinel-3 OLCI Level-2 product, \cite{S3OCLIdataguide}, which enables comparison with established methods. The first is for open waters and is a semi-analytical model based on in-situ data \cite{MOREL200769,olcichl}, while the second is a neural network (NN) based approach for complex waters. Suspended material significantly impacts the measurement of chlorophyll \cite{Siegel2005} and the NN based approach derives a series of inherent optical parameters from the reflectance spectra, which are then used to calculate chlorophyll-a concentration \cite{olcichl}.

CO2 fixation by marine phytoplankton accounts for about half the Earth’s primary production, and therefore plays a critical role in global climate change. Understanding what controls the trends and variability in the ocean carbon sink is a major question in Earth Science. Recent work from the Global Carbon Project suggests that model estimates of this sink are not in good agreement with observational-based evidence \cite{friedlingstein2022global}, but these estimates are needed by Earth System models used to make predictions of future climate change. Most satellite-based primary production models calculate daily water column production as a function of some measure of phytoplankton biomass and the photosynthetic response of phytoplankton to light \cite{WESTBERRY2023}. However, these models are recognised by the  Intergovernmental Panel on Climate Change (IPCC) to under-perform in estimating global prediction of primary production, due to lack of sufficient observational data and independent validation \cite{Bindoff2019}, with different methods often showing substantial discrepancies in estimated values \cite{Regaudie2014}. Data-driven methods offer the potential to address this by directly fitting to in-situ observations, but such approaches are even more severely limited by data availability. Due in part to the sparsity of data, studies have tended to use classical machine learning methods, for example, support vector machines \cite{Tang2008} and random forests \cite{ye2025,Zhang2025,ping2023}, using features derived from remote sensing data. While these machine learning models have performed well in specific locations compared to other models of primary productivity estimation, they have limited generalisability across different ocean ecosystems \cite{Zhang2025}.

Typically, foundation models are evaluated on a range of standardised datasets so their performance can be compared to other foundation models. This is useful in gauging relative performance, however in our case this is not possible, as no standard AI benchmarks exist for ocean colour applications and there are few models to compare to. Instead, we focus on evaluation for the two tasks described above, where in-situ data was readily available. We not only gauge the performance of fine-tuned models relative to the testing subset of the fine-tuning data, but also perform inference over large regions, in locations where there are no fine-tuning data. This allows us to compare the fine-tuned models with alternatives, including a model trained from `scratch', which consist of the same network architecture as the foundation model, but with randomly initialised weights and other physical and machine learning models. These experiments make it possible to understand how well this approach will perform in a wider range of situations and in real-world applications. 


The rest of this paper is organised as follows. Section \ref{sec:Data} details the datasets used for pre-training and fine-tuning. Section \ref{sec:methods} introduces the key components and steps of the methods. Section \ref{sec:results} presents the experiment results with a focus on fine-tuning experiments. Section \ref{sec:Discussion} discusses the key design considerations. 
Section \ref{sec:Summary} concludes this paper.

\section{Data} \label{sec:Data}

\subsection{Pre-training Data}

We use the Sentinel-3 A and B L2 OLCI full resolution water reflectances and the Sentinel-3 L2 SLSTR sea surface temperature as our primary datasets in this study. The Sentinel-3 OLCI sensor is a visible imaging push-broom radiometer and measures 21 bands ranging from 400 to 1200 nm.  We used 16 of the 21 bands (OL1 to OL12, OL16, OL17, OL18 and  OL21) in the pre-training, and excluded any bands used for only land or atmospheric measurements. The Sentinel-3 SLSTR sensor is a microwave radiometer and here we used the sea surface temperature band (SST). At the exception of the cloud flag value, we masked the OLCI/SLSTR measurements that were labeled with the flag values recommended by the Sentinel-3 product notice  \cite{OLCIflags}. We only included data that was at least 80$\%$ cloud free and this ensured that the quality of data in the pre-training step met the requirement.

We split the data into $45 \times 45$ image patches with no overlaps and resampled the SST temperature data to the same grid as the OLCI. Note that that we used a $42 \times 42$ image size for the pre-training and the larger image size allowed us to perform random cropping during fine-tuning (see Section \ref{sec:pretraining}). Even though both the OLCI and the SLSTR sensors are on the same satellite, they do not have the same look angles and this results in different parts of the image being affected by clouds. Therefore some parts of the image may have only water reflectances or sea surface temperature data. 

To obtain representative sampling, we collected an equal number of tiles (N=6400) from each Longhurst region \cite{longhurst2010ecological}. These regions split oceans into 83 biogeographic provinces. For this study we excluded the polar regions (APLR and ANTA) and inland water (LAKE). We used data between 2017 and 2021 and ensured an equal sampling thought the years by randomly sampling the same number of patches from each month. In total we collected 470,000 samples for training and 50,000 for validation. The data density is shown in Figure \ref{fig_datamap}, and by design the sampling strategy results in more data being used near the coast as there is greater diversity in data along the coasts compared the open ocean.

\subsection{Fine-tuning Data\label{sec:finetuning_data}} 

The chlorophyll and primary production datasets consist of single measurements for a particular time and location. To create fine-tuning datasets we collected cloud-free Sentinel-3 OLCI and SLSTR (using the same flags as in the pre-training) data over a 6 day window centered on the measurement. We then geo-referenced them to the same $80 \times 80$ grid and took the median over each pixel. The in-situ measurements were applied over roughly a one square km area so we labeled the six closest pixels and set the remaining pixels in the $80 \times 80$ label image as no-data. We chose an $80 \times 80$ image size to allow for larger cropping of the image during fine-tuning as discussed in \ref{finetuning}. 

\begin{figure}
    \centering 
    \includegraphics[width=1  \textwidth]{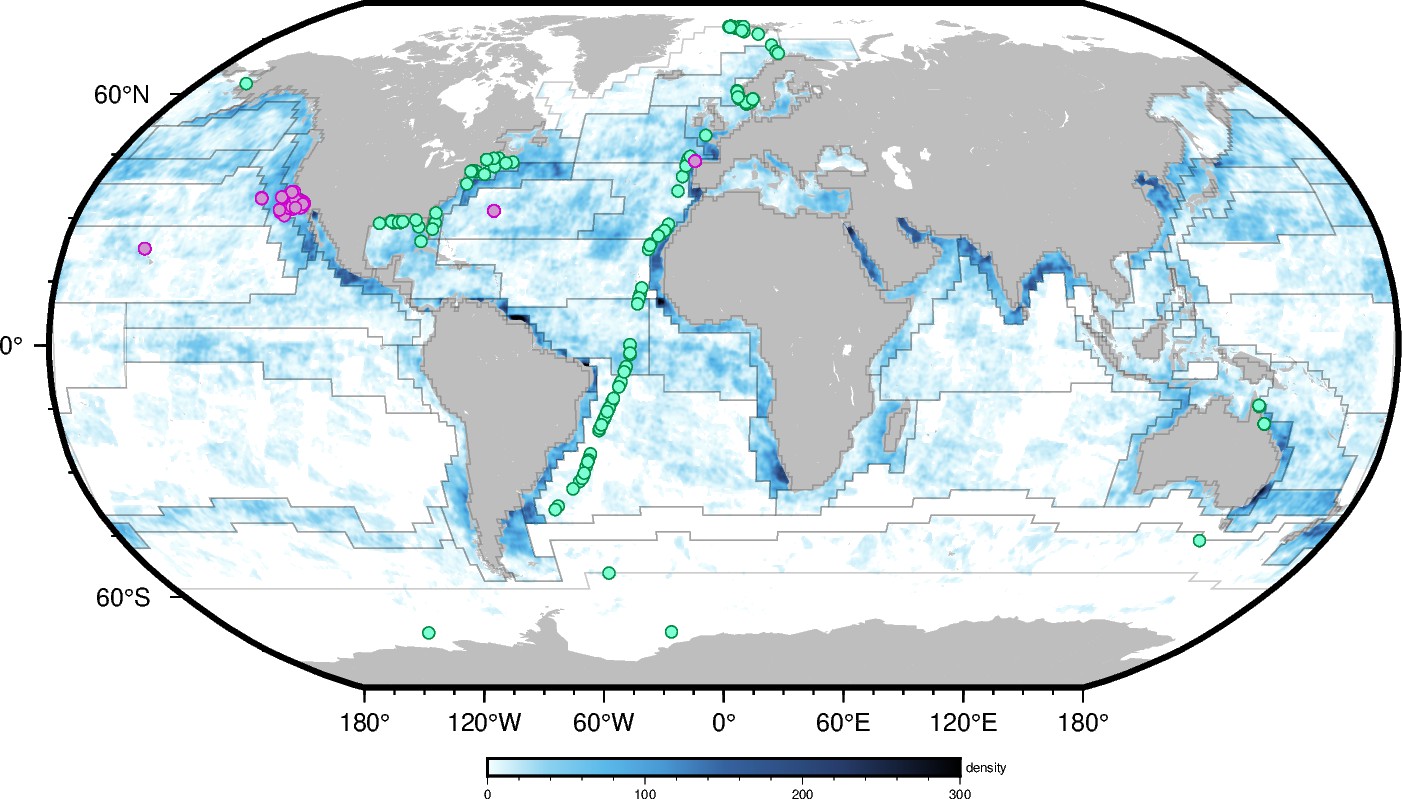}	
    \caption{Pre-training and fine-tuning data locations. The blue colour scale shows the density of pre-training data. The points in pink indicate the locations of in-situ primary production and chlorophyll observations are shown in green. Longhurst regions of the ocean are delineated in grey.} 
    \label{fig_datamap}%
\end{figure}

We used chlorophyll measurements taken from high performance liquid chromatography (HPLC) \cite{valente2022acog}. This dataset is a global compilation of in-situ data; a list of the original sources can be found in \cite{Valente2022}. In total we used 188 samples and these were mainly constrained to the Atlantic Ocean. The distribution of measurement locations can be seen in Figure \ref{fig_datamap}. We did not include SST in this dataset, as it is not typically used to derive chlorophyll concentrations \cite{TILSTONE2021}.

The primary production data were collected from various sources \cite{mattei2021, simonscmap202310019979, Marra2021,Buitenhuis2013,Goericke2021} and include ship-borne observations and buoy data. The locations can be seen in Figure \ref{fig_datamap}. Most data is either in the off the coastline of Spain and Portugal (N=14), off the West coastline of the USA (N=43). There are few points in the open ocean (N=46). In total there were N=103 measurements. 

For each of our in-situ sources, primary production measurements were taken at different sea depths. In order to take this variability into account while putting all of our primary production measurements into the reference system, we chose to compute the integrated sea depth version of these measurements using the trapezoidal rule. We acknowledge this introduces some uncertainty as the OLCI signal samples the data at an unknown, variable depth. However, initial experiments showed us that the `surface' (highest) measurements introduced even more variability into our model, while choosing a single depth would drastically reduce the size of our dataset while also not matching the depth used for sampling by the OLCI instrument.

\section{Methods} \label{sec:methods}

Our geospatial foundation model approach uses a backbone that is first pre-trained on a large amount of unlabeled data using a self supervised learning approach. This backbone is then combined with various heads (for example, a pixel-wise regression head, or a segmentation head) for fine-tuning on downstream application datasets. Our pre-training methodology is explained in Section \ref{sec:pretraining}, and our fine-tuning methodology is explained in Section \ref{finetuning}.

\subsection{Pre-training} \label{sec:pretraining}

The pre-training process of our foundation model follows the structure of the Prithvi-EO foundation models \cite{jakubik2023, szwarcman2024}. For the self-supervised pre-training, we use a masked autoencoder (MAE) approach: the goal is to reconstruct input images that are split into equal-sized, non-overlapping patches, with a percentage of patches randomly masked (the percentage of masked patches is a hyper-parameter set during validation). We use a Vision Transformer (ViT) architecture similar to \cite{szwarcman2024} for this task: the patches are flattened into token embeddings, and a linear projection is applied to the resulting tokens with additional 2D sin/cos positional embeddings. The unmasked tokens are then processed by the series of Transformer blocks of the encoder. The decoder then receives all of the encoder outputs along with placeholder tokens for the masked patches. Together, these tokens are then processed through a series of transformer blocks to reconstruct the image. The loss is calculated as the root mean square error (RMSE) between the reconstructed and original pixel values of the masked patches. 

The Prithvi-EO foundation models were trained using the Harmonized Landsat and Sentinel-2 (HLS) collection. These data have a grid spacing of 30m. With Prithvi-EO-2.0, the training images consisted of a stack of 4 HLS images of the same location captured at different times, giving a $4\times 224\times 224$ (stack size, image width, image height) image size. These images were then split into patches of size $4\times 16\times 16$ for the MAE pre-training. In contrast, for our application, the OLCI/SLSTR grid spacing is larger (300~m), and images over the ocean tend to be less complex. Also compared to land, what we observe is less dependent on exact positioning, as features over land tend to stay in the same place, hence stacking images multiple times over water is less meaningful. Therefore, we use a smaller image size of $1\times 42\times 42$ with a $1\times 2\times 2$ patch size. This has the advantage of finer pixel level processing, compared to using a larger patch size. Our model has 50 million parameters which is smaller than most models used in land applications. We found during pre-training and fine-tuning that a larger model did not have a demonstrable better performance, hence we used this smaller model. During training we also apply random crop and rotation to the input to get from the original $1\times 45\times 45$ dataset image size to the $1 \times 42\times 42$ input image size.

\begin{figure}
	\centering 
	\includegraphics[width=1  \textwidth,]{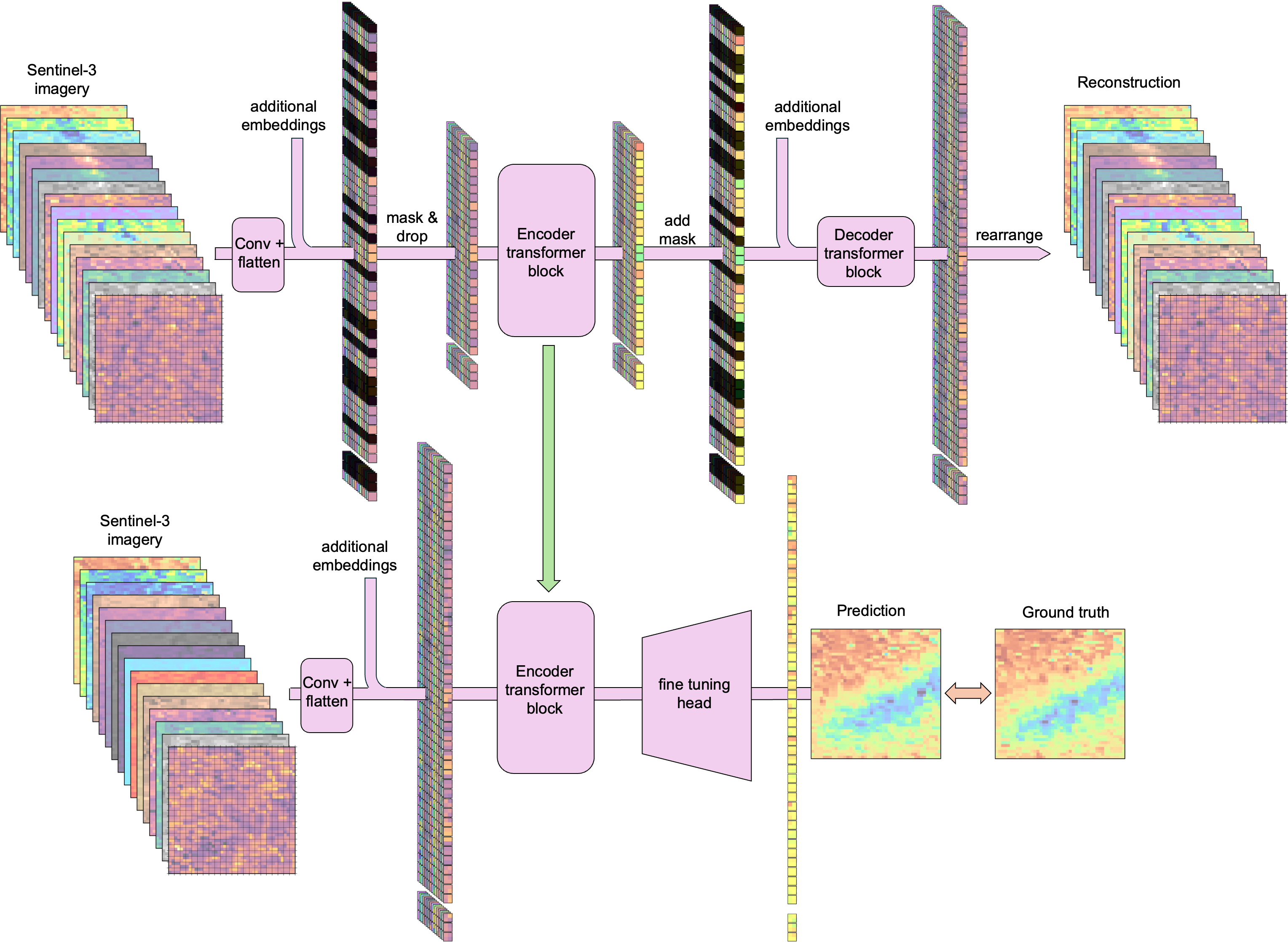}	
	\caption{Pre-training and fine-tuning architecture for the ocean color foundation model} 
	\label{fig_mom0}%
\end{figure}

\subsection{Fine-tuning} \label{finetuning}
As explained in Section \ref{sec:finetuning_data}, the foundation model is tested on two downstream tasks: primary production and chlorophyll quantification. To fine-tune for these tasks, we use the pre-trained encoder architecture with its best encoder weights obtained from the pre-training reconstruction task as described in Section \ref{sec:pretraining}. We then append a U-Net based decoder (\cite{xiao2018}, also see \cite{szwarcman2024} for the implementation), with a task-specific head for pixel-wise regression for both fine-tuning tasks. These networks process input data similar to the data that was used for pre-training: $42\times 42$ image crops with the same bands than the ones used for pre-training. 



\section{Results\label{sec:results}}

\subsection{Pre-training results}

We pre-trained two models: one with OLCI bands only, and one with OLCI bands and SST. We used a maximum learning rate of $2.4 \times 10^{-3}$ and trained using 4 Nvidia A100 80GiB GPUs, resulting in a training time of approximately 7-8 minutes per epoch. We ran pre-training over 100 epochs resulting in 11-12 GPU hours for full training. 

\begin{figure}
	\centering 
	\includegraphics[width=1  \textwidth,]{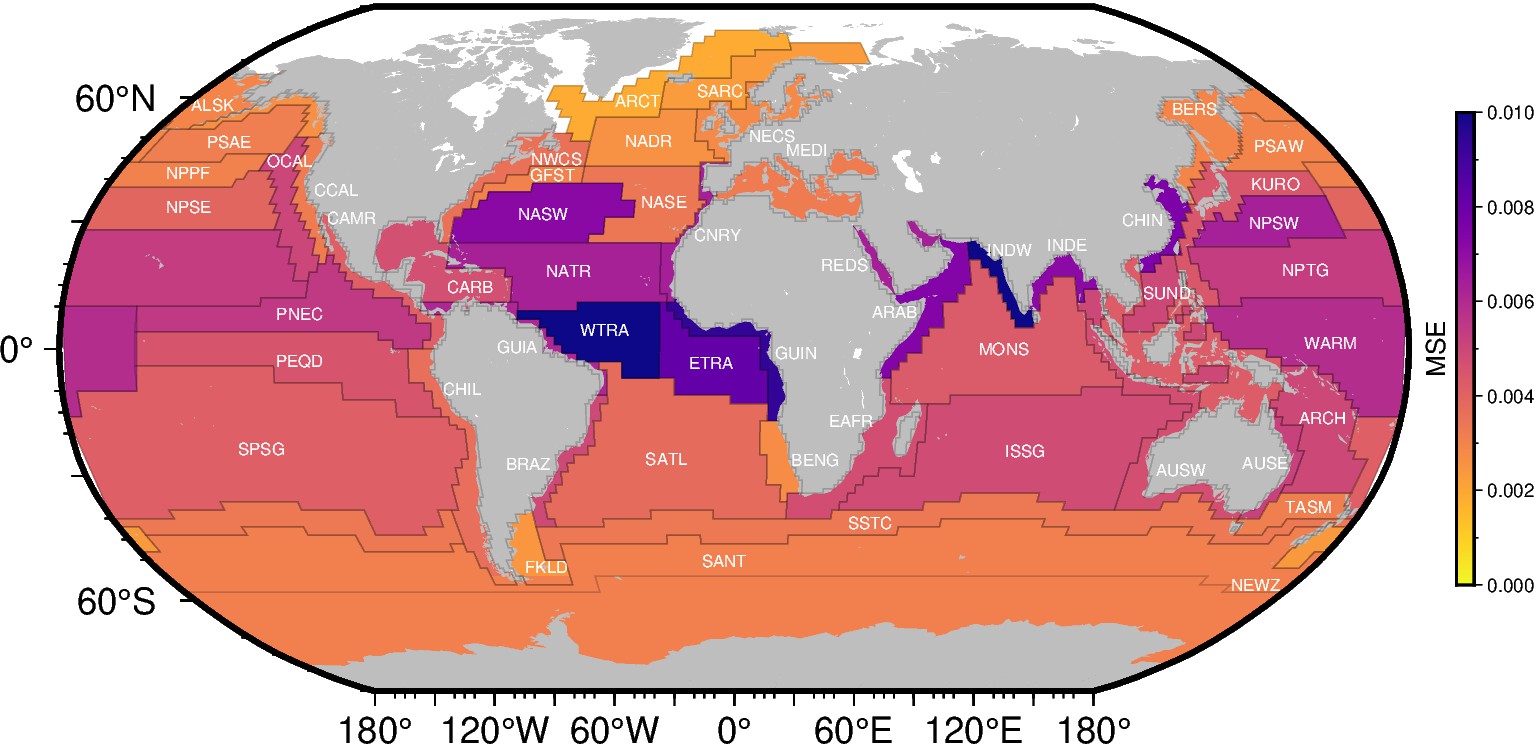}	
	\caption{Reconstruction loss by Longhurst region} 
	\label{fig_lprm}%
\end{figure}

The reconstruction loss per Longhurst region on the validation data is shown in Figure \ref{fig_lprm}. We observe that losses tend to be larger around the coastline and in the mid-Atlantic. Reconstruction losses are smaller when we include SST (see Table \ref{tab:results}).

\begin{table}
    \centering
    \begin{tabular}{|l|l|l|l|}
        \hline
        Model                            &  Reconstruction  & Chlorophyll      & Primary   \\
                                         &                  &   concentration &  production \\
                                         &                  &    $log_{10}[mg/m^3]$ & $log_{10}[mgC/m^2/day]$   \\
        \hline
        OLCI + (\textit{Scratch})                 &                  & $0.16\pm 0.03$   & $0.43\pm 0.05$ \\
        OLCI + (FM)                      & $0.0046\pm0.002$ & $0.16\pm 0.06$   & $0.39\pm 0.07$  \\
        OLCI + SST (\textit{Scratch})             &                  & $0.16\pm 0.10$   & $0.42\pm 0.07$ \\
        OLCI + SST (FM)                  & $0.0047\pm0.002$ & $0.14\pm 0.05$   & $0.39\pm 0.04$  \\
        Random forest                    &                  & $0.16\pm 0.10$   &  $0.40 \pm 0.04$\\
        \hline
    \end{tabular}
    \caption{RMSE cross validation losses (average and standard deviation) for the reconstruction, chlorophyll concentration and primary production tasks (see Section \ref{sec:methods} for more details). The pre-trained model is either the ocean colour only model (OCLI), or the ocean colour and SST model (OLCI + SST). For the downstream tasks, the model is either the foundation model (FM), the model trained starting from random weights (\textit{scratch}), or a random forest model.}
    \label{tab:results}
\end{table}

\begin{figure}
	\centering 
	\includegraphics[width=1  \textwidth,]{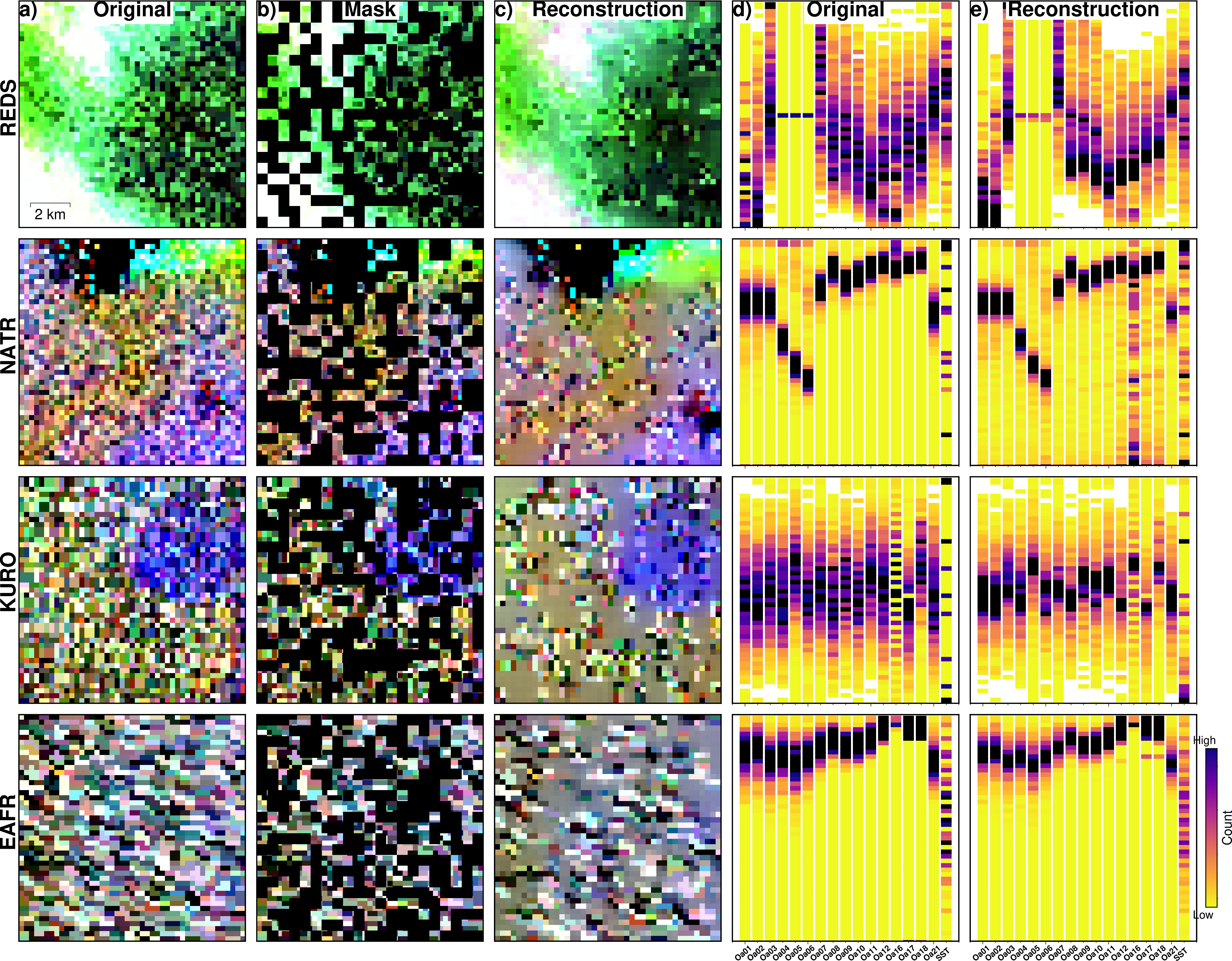}	
	\caption{Pre-training reconstructions, with column a) being the original image b) the mask, c) the reconstruction. The images are shown in the `Tristimulus' band recipe \cite{learnolci2025}. Columns d) and e) show the normilised distribution of reflectances for each band for the original and reconstructed image respectively. } 
	\label{fig_recon}%
\end{figure}

The original images, masks and reconstructions of four image patches from different regions (EAFR, KURO, NATR and REDS) are shown in Figure \ref{fig_recon}, along with the normalised histograms for each band in the image, including the SST. Both the reconstructed images and histograms show the model performs well in reconstructing the data. The reconstructed images are smoother than the inputs, which is by design, as we deliberately use a small decoder which will be discarded when fine-tuning for downstream tasks.

\subsection{Fine-tuning results}

We performed fine-tuning using Terratorch \cite{IBMTerratorch2025}, a fine-tuning toolkit for geospatial foundation models that uses PyTorch lightning and Torch-Geo. 


In order to assess the value of using the geospatial foundation model approach, we compared the fine-tuned results from our two pre-trained models to a \textit{scratch} model, consisting of the same network architecture as the foundation model, but with randomly initialised weights, and a random forest model (the ExtraTreesRegressor from \cite{scikit-learn}). We also performed a series of unique experiments for each task. For the chlorophyll task, we focused on comparing our results to other models, including the Level-2 OLCI product, and for the primary production task we also investigated the influence of the number of samples used in fine-tuning.

As explained in Section \ref{sec:finetuning_data}, our fine-tuning datasets consist of 188 and 103 observations for the chlorophyll and primary production tasks respectively. Each observation is an image patch of size $80\times 80$. Its mask is also a $80\times 80$ image where the label corresponding to the numerical value of the target quantity (e.g. integrated phytoplankton production for the primary production task) is a float value assigned to a $3\times 3$ location. The rest of the mask remains unlabeled. Data augmentation is performed during training: first, similar to the pre-training, a random crop of a $42\times 42$ sized patch, followed by a random application  of vertical and horizontal flips (50$\%$ chance each), followed by a random rotation ranging from -30$^{\circ}$ to 30$^{\circ}$. Also we trained with the $log_{10}$ of both the chlorophyll-a concentration and primary production as both values typically have a large range of values. 

Due to the small amount of observations in our fine-tuning datasets, we chose to evaluate the performance of our models using 5-cross fold validation: for each set of hyperparameters, we fine-tuned (foundation models) / trained (scratch models) our models on each of the 5 training folds, and evaluated their performance over the corresponding validation fold. Note that the folds were randomly created without taking into account the origin (source or location) of the observations.

\subsubsection{Chlorophyll concentration task}

The results for the 5-cross fold validation experiments (Table \ref{tab:results}) show that the RMSE testing loss is similar for all models including ones that has been trained from scratch, used pre-trained weights or that that includes SST (in this case as we are not including SST in the fine-tuning data those weights will be ignored). Additionally, losses for in the FM-based models were similar to those of the decision tree model.

\paragraph{Large scale inference}

\begin{figure}
	\centering 
	\includegraphics[width=1  \textwidth,]{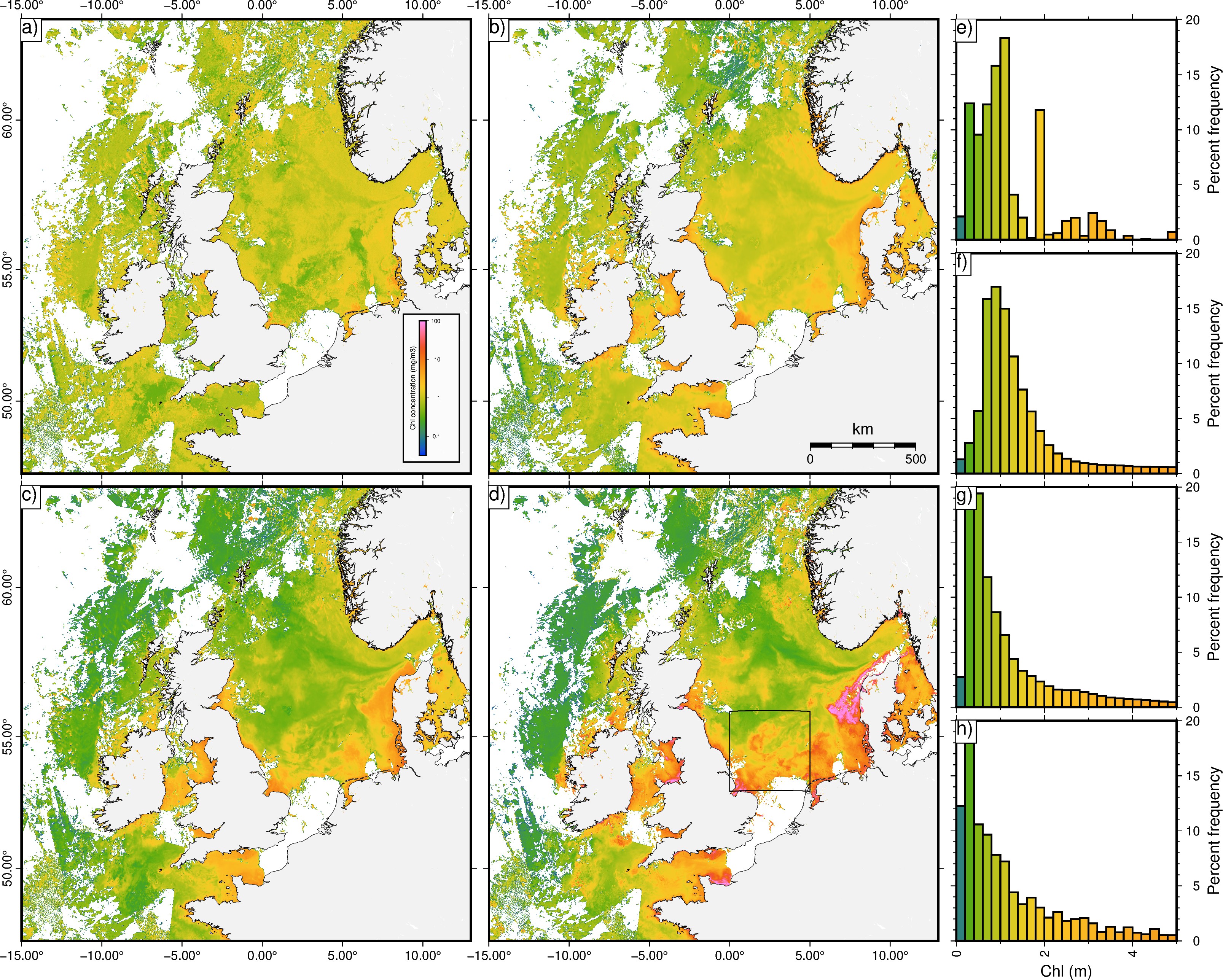}	
	\caption{Chlorophyll-a concentration ($mg/m^3$) spatial patterns (a-d) and distributions (e-h) predicted by: (a,e) decision tree (b,f) \textit{scratch} model (c,g) fine-tuned foundation model and (d,h) L2 NN model. The box in d) is the region of interest in Figure  \ref{fig_chl_zoom} .} 
	\label{fig_chl}%
\end{figure}

\begin{figure}
	\centering 
	\includegraphics[width=1  \textwidth,]{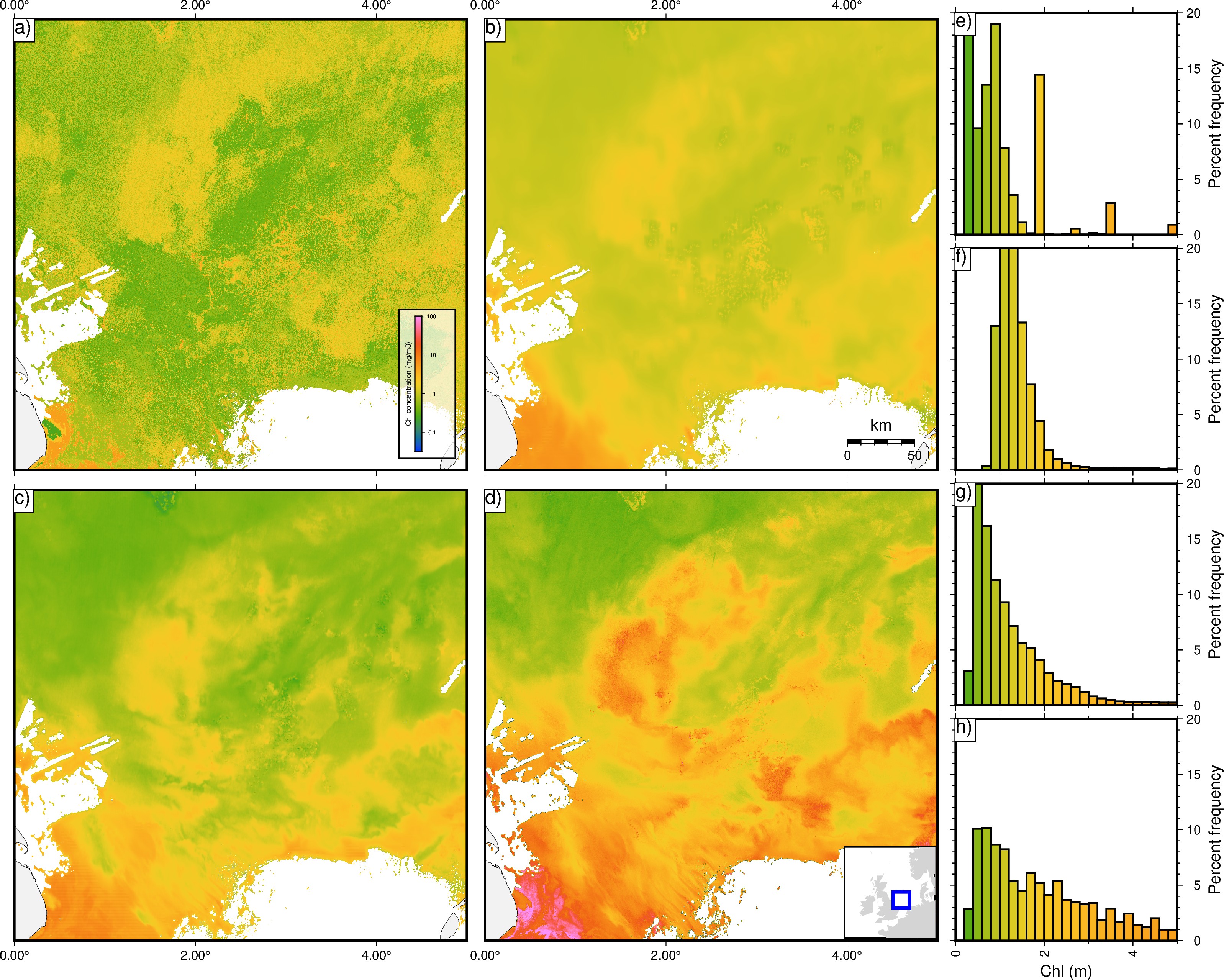}	
	\caption{Sub region chlorophyll-a concentration ($mg/m^3$) spatial patterns (a-d) and distributions (e-h) predicted by: (a,e) decision tree (b,f) \textit{scratch} model (c,g) fine-tuned foundation model and (d,h) L2 NN model} 
	\label{fig_chl_zoom}%
\end{figure}

As explained in Section \ref{sec:finetuning_data}, our dataset measurements are scarce, with only at best a $3\times 3$ labeled area per $42\times 42$ patch and therefore we are sampling a relatively small geographical domain. To test how well the models generalise spatially, we conducted a second experiment, performing inference over a large region around the North Sea (see Figure \ref{fig_chl}, and used the Level-2 NN model for the comparison (as the majority of the area is coastal). All models trained using the chlorophyll-a in-situ measurements do not measure values above ~10 $mg/m^3$ (which are present in Level-2 NN based model) this is due to the fine-tuning data having a maximum concentration of chlorophyll-a of 8.85 $mg/m^3$. Despite this, we observe that the model fine-tuned from pre-trained weights more closely matches the spatial patterns and the distribution of concentrations of Level-2 NN based model (Figure \ref{fig_chl}). In particular, the model trained from scratch is smoother compared to both the fine-tuned model and the Level-2 NN model, while the decision tree model displays features that to not match any other model. Further analysis reveals that the fine-tuned models has a larger structural similarity index (a measure of the perceived quality of the image), of 0.88, compare to the decision tree (0.68) and the model trained from scratch (0.82). This can be seen in more detail when by inspecting a smaller area (Figure \ref{fig_chl_zoom}): the fine-tuned model is the only model to match the spatial pattern of OLCI Level-2 NN based model correctly.

\subsubsection{Carbon primary production task.}

The best 5-cross fold validation results for primary production are presented in Table \ref{tab:results}. For both the OLCI and the OLCI+SST model, we can see that using the foundation model pre-training allows us to improve the average RMSE loss. The best-performing model is the fine-tuned model without SST, with a decrease in loss of 0.05 (11.8\%) compared to the \textit{scratch} model. 

\paragraph{Large scale inference}

Again, in order to test how well the model could generalise spatially, we conducted a second experiment, performing inference over a large region that was not in our training/validation dataset. Because in-situ primary production measurements are not generally or openly available for other regions, we can only evaluate the ability of the FM-derived models to replicate spatial primary production patterns commonly seen with other primary production models.

\begin{figure}
    \centering
:    \includegraphics[width=1\linewidth]{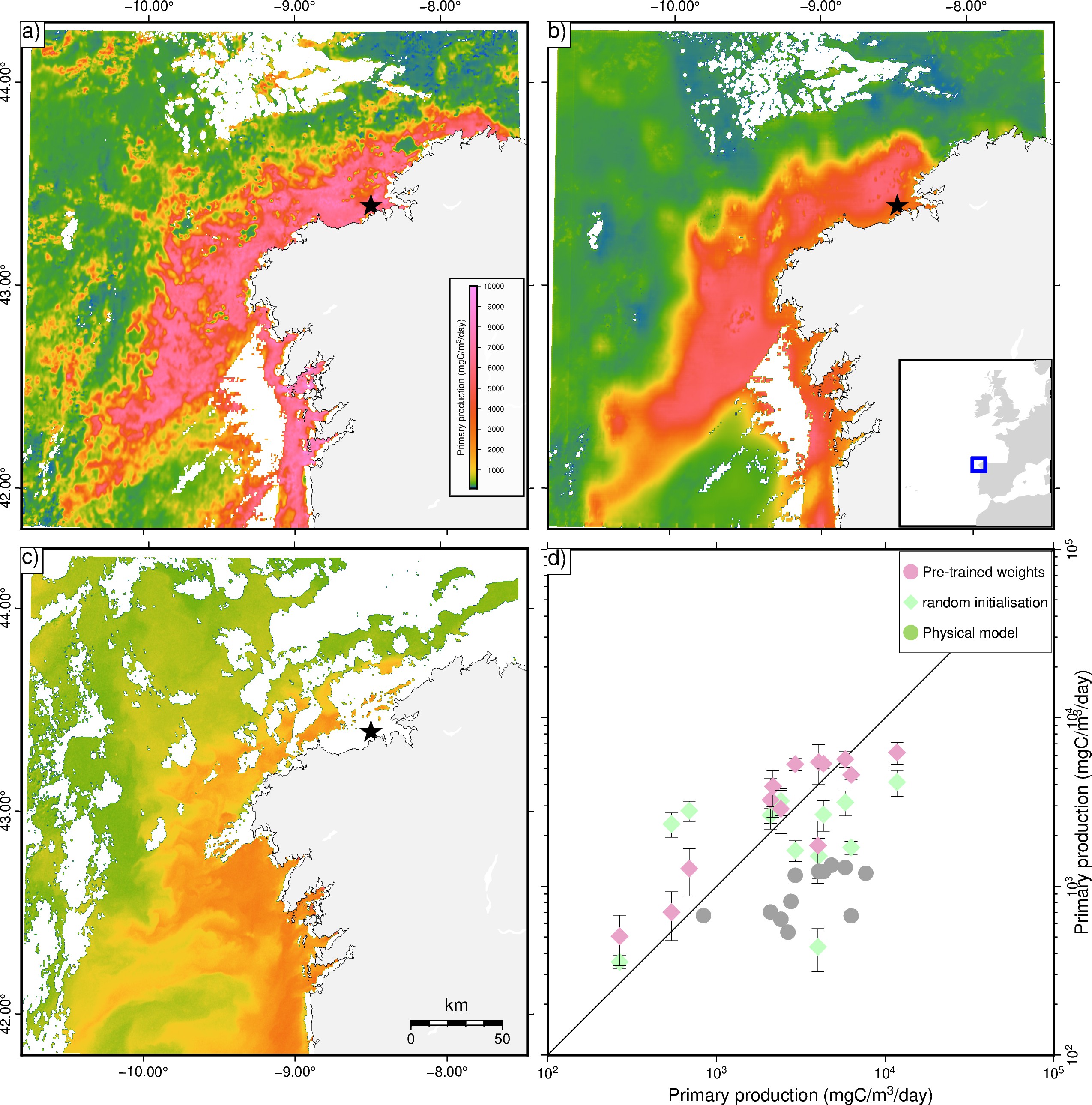}
    \caption{Primary production shown on log scale. (a) Inference from model fine-tuned on OCLI+SST; (b) inference with \textit{scratch} model; (c) prediction of physical model; (d) comparison of models at measurement site (location indicated by a star on the spatial plots) }
    \label{fig:bofb_val}
\end{figure}

A comparison between the 7th and 13th July 2020 for the coast of Spain and Portugal is shown in Figure \ref{fig:bofb_val}. This region is known to have high levels of primary production, and tends to be a region where models traditionally perform poorly. We can first observe that the models compare well to the physical model, and produce a realistic primary production pattern. The second observation that we can make is that the output of the models trained from scratch is smoother and less spatially precise than the fine-tuned model, a feature which is not apparent from inspecting only the cross-validation results of Table \ref{tab:results}. This region also contains one in-situ measurement site (held out from the training). We observe that the fine-tuned foundation model more closely matches the values at this site compared to the model trained from scratch, while the physical model consistently under predicts the primary production (Figure \ref{fig:bofb_val}d).


\paragraph{Influence of the number of dataset samples used}

With this third experiment on the primary production task, we evaluated the impact of fine-tuning the foundation model using smaller datasets. For this, we repeated the cross validation experiment for the OLCI+SST model, and used the same fold splits. However, for each training fold, we limited the percentage of available observations ($12.5, 25, 37.5, 50, 62.5, 75, 87.5$ and $100\%$).

\begin{figure}
    \centering
    \includegraphics[width=1\linewidth]{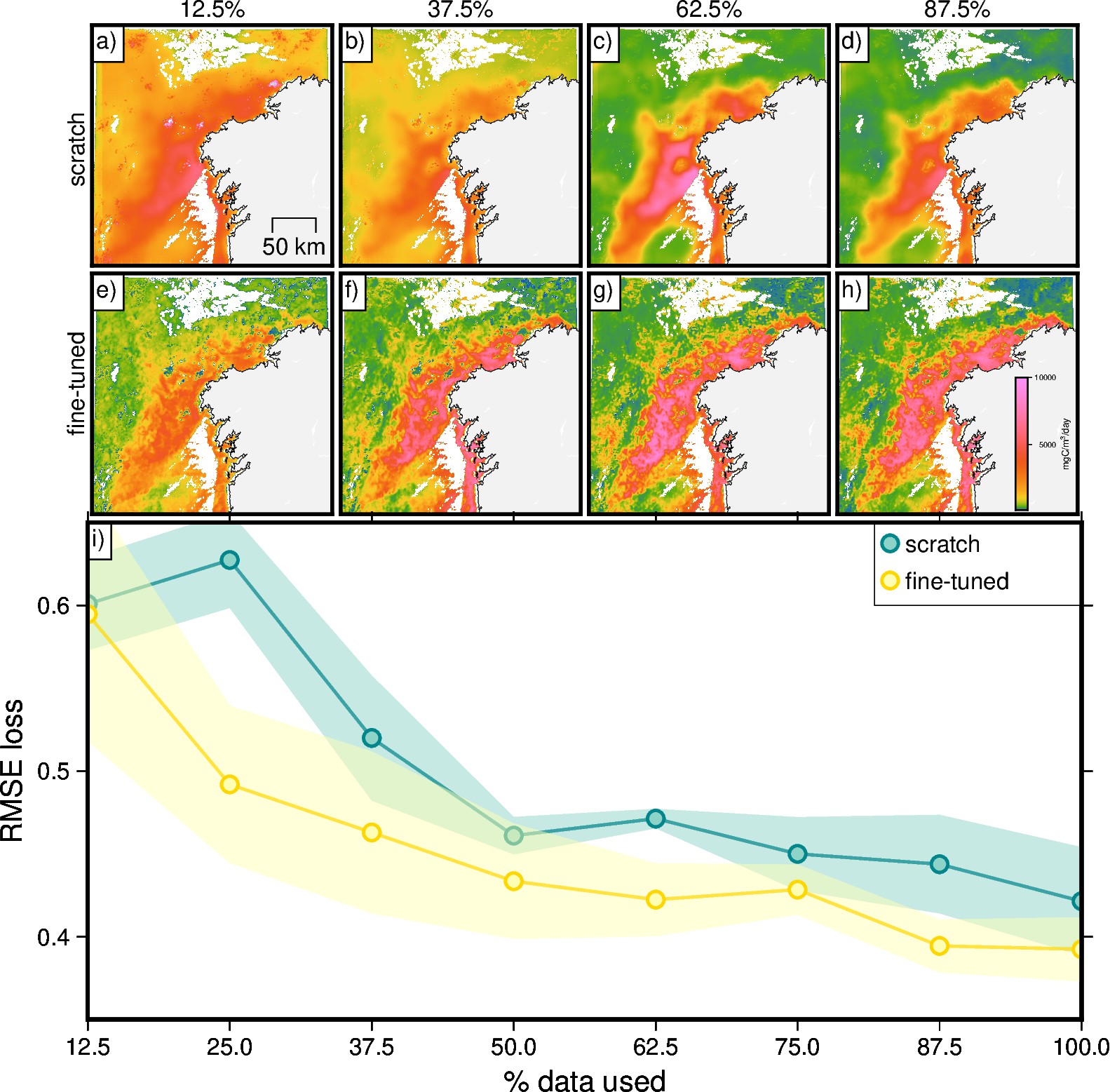}
    \caption{Spatial patterns and RMSE loss of primary production models using limited percentage of the training dataset}
    \label{fig:pctg_data_pp}
\end{figure}

The results are reported in Figure \ref{fig:pctg_data_pp}. At $12.5\%$ of available dataset, both approaches have similar results, suggesting that their performance is close to random. At $25\%$, while the \textit{scratch} model performance gets slightly worse, the fine-tuned model performance improves significantly, suggesting it was able to learn to detect phytoplankton production with only $25\%$ of the training datasets (19 observations). The difference between the fine-tuned model and the model trained from scratch reduces for larger amounts of data, but the fine-tuned model continues to consistently outperform the model trained from scratch.


\section{Discussion} \label{sec:Discussion}

Both fine-tuned models perform well with very little data, with the chlorophyll-a and the primary production models fine-tuned with scarce labels totaling only 7\% and 6\% of the number of pixels in a single image respectively, demonstrating the advantage of the foundation models approach in terms of data efficiency compared to other methods. For larger fine-tuning datasets, there may be a smaller performance advantage in the foundation model approach, as illustrated by the trend shown in Figure \ref{fig:pctg_data_pp}. 

Our results also illustrate that, for this class of task, the difference in performance between models are not always reflected in summary statistics from the training, in part due to the small number of testing samples. For example, the RMSE from chlorophyll-a concentrations task did not indicate well how well the spatial patterns compared to the standard Level-2 product. It was only by examining the spatial plots that we could observe the substantial difference in detail between the fine-tuned foundation model and the model trained from scratch. A likely explanation of this is due to the nature of our training images and labels: the models are trained and optimised on a dataset of $42\times 42$ crops with only a $3\times 3$ set of values to consider, the rest of the image being ignored. These results show that the ability of the foundation model to create detailed spatial patterns which match the local pixel observations, while the \textit{scratch} model tends to produce an average value across the same extent as a training image. This effect is even more apparent in the primary productivity task where there is less training data.

Even though the foundation model approach works well, our results also illustrate that fine-tuned models still ideally need to be provided with a range of data representing the entire parameter space in order to be fully generalisable.
Since the chlorophyll-a data did not include measurements of very high concentrations, the model under-predicts these for unseen locations. In contrast, for the primary production task, where high values were present in the fine-tuning data, the model was able to predict higher primary productivity near the coast compared to a physical model, although validation against observations was again limited by the dataset size.

Because both our fine-tuning datasets are small, the fine-tuned models created for this study do not perform well enough to be used as a replacement for other more well-established techniques. They instead serve as examples as how the ocean colour foundation model can be used, and the amount of fine-tuning data needed. We therefore provide the fine-tuned dataset for primary production, and the foundation model pre-trained weights, so they can be used as examples for other tasks. Fine-tuning of these models also does not require many resources: for example fine-tuning the primary production model approximately 12~minutes on a NVidia T4.

The model architecture presented here, based on the Prithvi-EO, has shown to perform well on a range of different tasks, and even though we have presented two regression tasks, we the ocean colour foundation model can also be fine tuned for classification, segmentation and object detection, given the relevant labelled data for fine-tuning.

\section{Summary and conclusions} \label{sec:Summary}

We created a foundation model base on Sentinel-3 data for ocean colour applications. This model was based on the Prithvi-EO foundation model architecture and was adapted for use with Sentinel-3 data. Specifically, we reduced the input image size from $224\times 224$ to $42 \times 42$ account for the larger pixel size (300 m) of Sentinel-3 OLCI data compared to the Prithvi-EO model which was trained on 30 m resolution data. The ocean foundation model was trained using 512,000 tiles, distributed geographically by using the same number of tiles in each Longhurst region, while also keeping an equal number of samples in each month for each region. We also reduced the model size to 50,000 parameters, as we observed no additional benefit in the pre-training to having a larger model. We also investigated adding sea surface temperature as an additional band from Sentinel-3 SLSTR.

We chose chlorophyll concentration quantification and carbon primary production quantification as our downstream tasks to test our model. For chlorophyll concentration, the performance statistics are very similar for both fine-tuned models (with and without SST) and those without pre-training. For the primary production model, the performance statistics did demonstrate an advantage in FM pre-training, particularly when the fine-tuning dataset size was further reduced. Visualisation of the spatial patterns produced by the models further demonstrated that the point-based evaluation of RMSE does not give a full picture of model performance: when we ran inference over a large, unseen area, we observed that the models based on FM pre-training produced qualitatively more realistic patterns. We note that the performance of the fine-tuned models are still dependent on the quality of the input data and will not be able to generalise well outside the range of values and parameter spaces represented by those locations. However, we have demonstrated the power of foundation models in using very small amounts of representative samples to obtain good performance. 

\section*{Acknowledgements}
This work was supported by the Hartree National Centre for Digital Innovation, a collaboration between STFC and IBM. 



\end{document}